\documentclass[conference]{IEEEtran}
\usepackage{graphicx}
\usepackage{times}
\usepackage{epsfig}
\usepackage{graphicx}
\usepackage{amsmath}
\usepackage{amssymb}
\usepackage{cite}
\usepackage{subcaption}
\usepackage{bm}
\usepackage{tabularx}
\usepackage{xcolor}

\usepackage{amsmath,amssymb,amsthm}
\usepackage{algorithm,algorithmic} 

\begin{document}
%
\title{Unsupervised Temporal Feature Aggregation for Event Detection in Unstructured Sports Videos}

\author{\IEEEauthorblockN{Subhajit Chaudhury}
\IEEEauthorblockA{IBM Research AI\\
Email: subhajit@jp.ibm.com}
\and
\IEEEauthorblockN{Daiki Kimura}
\IEEEauthorblockA{IBM Research AI\\
	Email: daiki@jp.ibm.com}
\and
\IEEEauthorblockN{Phongtharin Vinayavekhin}
\IEEEauthorblockA{IBM Research AI\\
	Email: pvmilk@jp.ibm.com}
\and
\IEEEauthorblockN{Asim Munawar}
\IEEEauthorblockA{IBM Research AI\\
	Email: asim@jp.ibm.com}
\and
\IEEEauthorblockN{Ryuki Tachibana}
\IEEEauthorblockA{IBM Research AI\\
	Email: ryuki@jp.ibm.com}
\and
\IEEEauthorblockN{Koji Ito}
\IEEEauthorblockA{Japan Institute of Sports Sciences\\
	Email: koji.ito@jpnsport.go.jp}
\and
\IEEEauthorblockN{Yuki Inaba}
\IEEEauthorblockA{Japan Institute of Sports Sciences\\
	Email: yuki.inaba@jpnsport.go.jp}
\and
\IEEEauthorblockN{Minoru Matsumoto}
\IEEEauthorblockA{Japan Institute of Sports Sciences\\
	Email: minoru.matsumoto@jpnsport.go.jp}
\and
\IEEEauthorblockN{Shuji Kidokoro}
\IEEEauthorblockA{Japan Institute of Sports Sciences\\
	Email: shuji.kidokoro@jpnsport.go.jp}
\and
\IEEEauthorblockN{Hiroki Ozaki}
\IEEEauthorblockA{Japan Institute of Sports Sciences\\
	Email: hiroki.ozaki@jpnsport.go.jp}}

\maketitle

\begin{abstract}
	Image-based sports analytics enable automatic retrieval of key events in a game to speed up the analytics process for human experts. However, most existing methods focus on structured television broadcast video datasets with a straight and fixed camera having minimum variability in the capturing pose. In this paper, we study the case of event detection in sports videos for unstructured environments with arbitrary camera angles. The transition from structured to unstructured video analysis produces multiple challenges that we address in our paper. Specifically, we identify and solve two major problems: unsupervised identification of players in an unstructured setting and generalization of the trained models to pose variations due to arbitrary shooting angles. For the first problem, we propose a temporal feature aggregation algorithm using person re-identification features to obtain high player retrieval precision by boosting a weak heuristic scoring method. Additionally, we propose a data augmentation technique, based on multi-modal image translation model, to reduce bias in the appearance of training samples. Experimental evaluations show that our proposed method improves precision for player retrieval from 0.78 to \textbf{0.86} for obliquely angled videos. Additionally, we obtain an improvement in F1 score for rally detection in table tennis videos from 0.79 in case of global frame-level features to \textbf{0.89} using our proposed player-level features. Please see the supplementary video submission at https://ibm.biz/BdzeZA.
\end{abstract}
\begin{IEEEkeywords}
	Video Processing, Person Re-identification, Sports Analytics, Clustering, Generative Adversarial Networks, Deep Learning for Multimedia
\end{IEEEkeywords}

\section{Introduction}
Recent growth in sports-related content has enabled teams with a plethora of data for concise and robust sports analytics to improve their gameplan. However, analyzing an exhaustive list of such videos by manual inspection is an insurmountable task that would require an enormous amount of manpower and resources. A viable solution to this problem can be to automatically extract interesting points in the game as a preliminary stage by a machine learning system. 
Human analysts can then use the automatically extracted video excerpts to provide their expert views on selected portions of the video. 
For example, in the game of soccer, the task of detecting \emph{corner kicks} can be automated by an image-based sports analytics system while the more subtle details like player positioning, kick angle, opponent strength, etc. can be analyzed by humans, thus providing a speed up to the entire analytics process.

\begin{figure}[tb]
	\centering
	\includegraphics[width=0.5\textwidth]{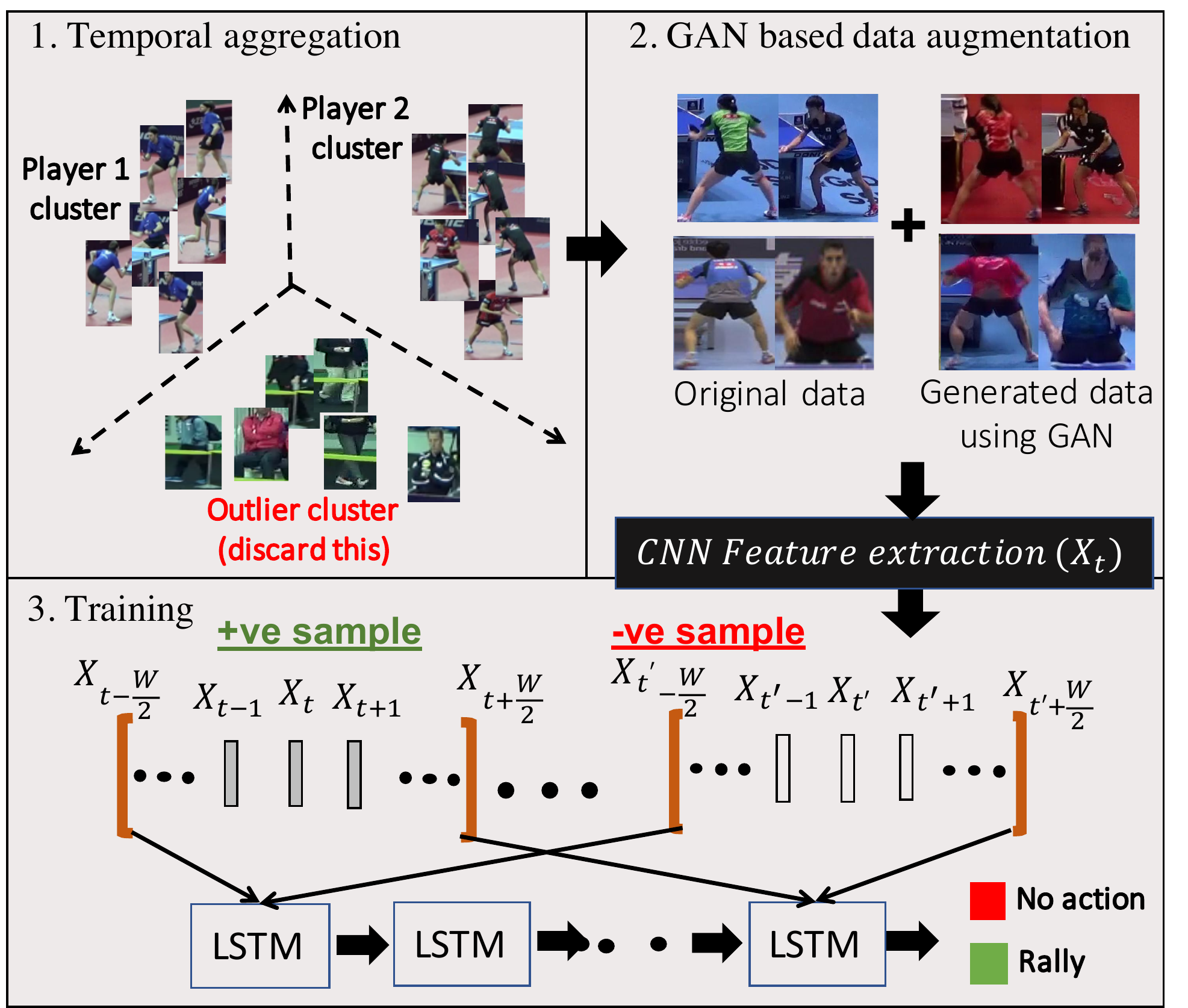}
	\caption{Overview of the proposed event detection system for table tennis videos. We perform a temporal feature aggregation and clustering-based player retrieval and data augmentation by using a multi-modal image translation model. Training is performed on the augmented data to learn a rally detector. }
	\label{fig:intro}
\end{figure}


With current breakthroughs in image classification~\cite{vgg, resnet, densenet}, object recognition~\cite{redmon2016you, fasterrcnn, ssd} and activity recognition~\cite{simonyan2014two,tran2015learning}, it is feasible to use such representation learning-based systems for auto-tagging and segmenting sports videos. There have been notable works in the direction of automated tagging in the sports videos \cite{yoshikawa2010automated, yan2014automatic, ghosh2018towards}, which uses motion-based segmentation models and activity recognition using deep learning features for key-point extraction in videos. While these methods provide useful tools for sports analytics, they tend to primarily focus on structured video settings, like TV broadcasts or special camera setup and assume that a large number of labels are available at training time. However, a major portion of sports-based contents on the Internet are unstructured, captured from arbitrary shooting angles. Thus, the existing methods are not suitable for handling such videos captured in-the-wild and our main focus in this paper is to study the challenges in analyzing such unstructured videos to develop a robust and accurate event detection system in such scenarios.

\begin{figure}[t]
	\centering
	\includegraphics[width=0.48\textwidth]{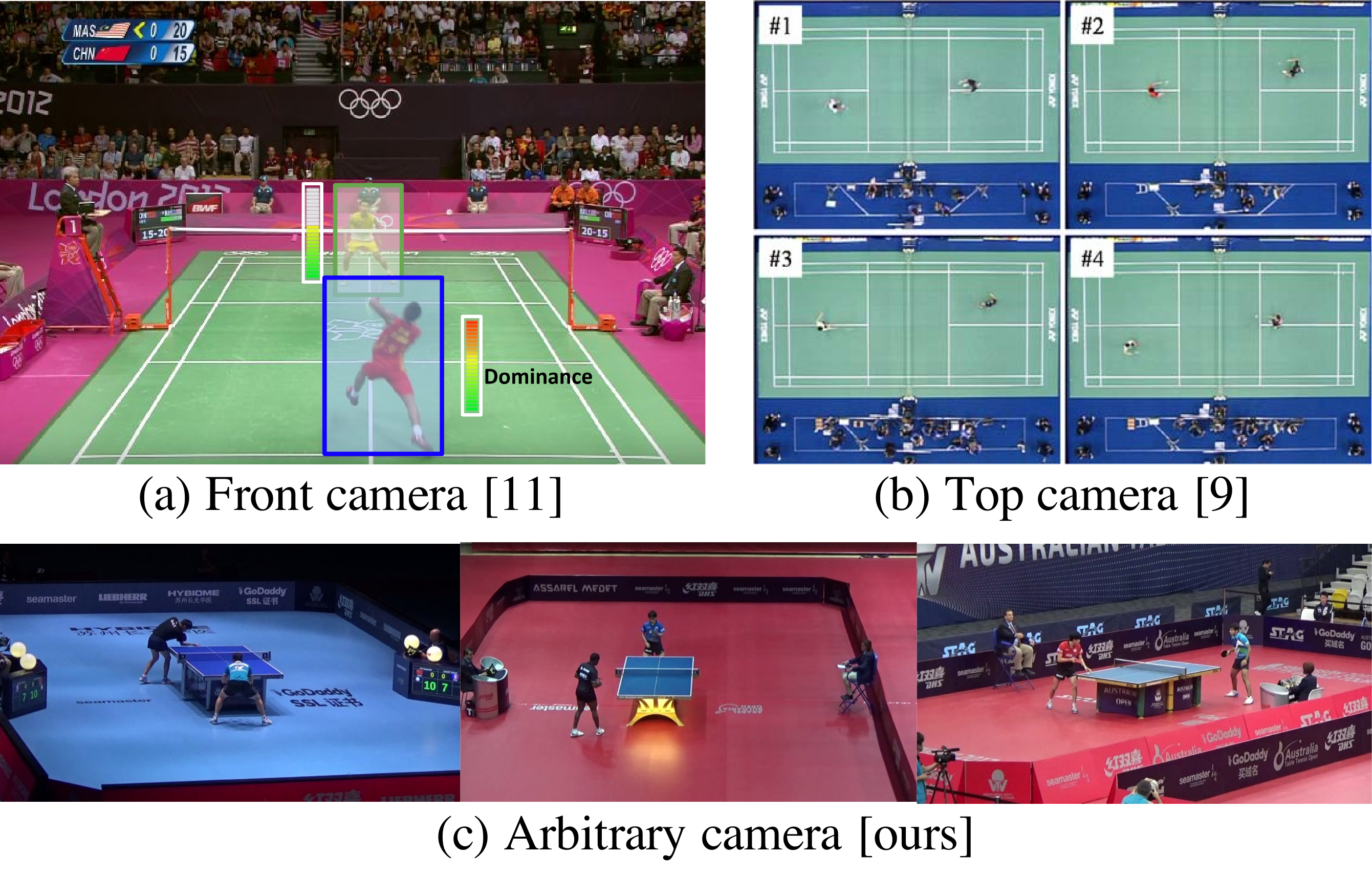}
	\caption{Our method handles arbitrary camera poses compared to structured settings like front or top camera in prior works. }
	\label{fig:compare}
\end{figure}

In this paper, we focus on the problem of rally detection in table tennis videos captured from arbitrary camera angles with real-time performance~($10 \sim 25$ FPS) to enable the human analysts with game summary soon after the game ends. Figure~\ref{fig:intro} shows an overview of our pipeline for rally detection. We train a detector model that uses player-level image features to classify whether a window of video frames belong to a rally event or not. We identify two major problems related to our application. Firstly, we do not have access to player bounding box annotations during training and hence we need to perform player retrieval from each image frame in an unsupervised manner without any user intervention. Secondly, our dataset has high variation in camera pose, but limited variations in appearance features such as color, which makes it liable to overfitting. Figure~\ref{fig:compare} shows the comparison between structured and unstructured settings of video capture for sports applications.

For the first problem, our method finds all ``person" bounding boxes in each frame using a pre-trained object detector. These are then ranked in the order of likelihood of being a player using a heuristic scoring mechanism. However, such a heuristic ranking method is weak and returns multiple noisy detections (like background audience, other players, coaches, etc.). 
To address this issue, we perform temporal feature aggregation to boost the performance of the weak heuristic score. The idea is that aggregation of person re-identification features of the cropped player bounding boxes from multiple temporally separated frames would form two dominant clusters in a latent representation space and proximity of candidate detections to these cluster centers can be used for boosting simple heuristics-based method. To address the second issue of preventing our model from overfitting to appearance features in the training set, we perform data augmentation using Multimodal UNsupervised Image Translation (MUNIT)~\cite{huang2018multimodal}. The translation model preserves contents of the training samples~(like player pose and actions) while changing the color distribution in the sample according to a randomly drawn style code, thus increasing variability in appearance-based features.

 Experimental evaluations indicate that our proposed temporal feature aggregation method can improve the precision of player retrieval. Similarly, our rally detector model trained using player-level features can improve F1 Score from 0.79 to 0.89 when compared to the baseline global frame-level feature-based detectors. In summary, we make the following 3 major contributions in this paper:
\begin{itemize}
	\item [1.] We propose a novel algorithm for heuristic score boosting using temporal feature aggregation based outlier rejection for player retrieval without user intervention.
	
	\item [2.] We propose a data augmentation method using an image translation model retaining the base content of the samples while increasing variability in appearance features to improve model generalization to unseen test cases.
	
	\item [3.] We present a real-time automatic rally scene detection method from unstructured table tennis videos shot at arbitrary camera angles.
	
\end{itemize}

\section{Related Works}
Image-based sports analysis has been extensively used for information extraction from various sports. Literature in this area can broadly be categorized into three approaches: (1) Using general action recognition methods, (2) Detecting a location of the ball or shuttle, (3) Analyzing the characteristic of the players. We provide the details of each area below,

\textbf{Using general action recognition}:
Methods in this category use techniques from action recognition literature \cite{simonyan2014two,tran2015learning,lea2016segmental} to detect important events in sports videos. Ramanathan~\emph{et al}.~\cite{Ramanathan_CVPR2016} proposed a model that learns to detect events and key actors in basketball games by tracking players using Recurrent Neural Networks~(RNN) and attention. Singh~\emph{et al}.~\cite{SINGH201745} tried to perform action recognition using feature trajectories of images from the first-person view camera. Another class of methods based on the ActivityNet~\cite{ghanem2018activitynet} dataset performs temporal segmentation as a sub-problem. However such methods use complex network architectures and multiple features like RGB+optical flow, which increases computational complexity at inference.  However, our application requires fast real-time performance allowing the coach or analyst to obtain the game summary as the game progresses. Thus such methods are not suitable in the scope of this application.

\textbf{Detecting the location of ball}:
Reno~\emph{et al}.~\cite{Reno_2018_CVPR_Workshops} used Convolutional Neural Network~(CNN) to detect the trajectory of a tennis ball in a video sequence with high accuracy. 
Dierickx~\cite{Tom_2014} detected a 3D trajectory of a shuttlecock to determine the different stroke types in badminton. Vučković~\emph{et al}.~\cite{Vuckovic_2014} assessed tactics in squash by considering ball locations when the shots are played. 
Tamaki~\emph{et al}.~\cite{Tamaki_2013_CVPR_Workshops} tried to detect the ball of table tennis by two RGB cameras or a single RGB-D camera with known corner position of the table and camera parameters, whereas in our problem setting we aim to detect events in single RGB video without such meta-information.

\textbf{Analyzing characteristics of players}:
Weeratunga~\emph{et al}.~\cite{Weeratunga_2017_CVPR_Workshops} proposed a method to analyze badminton tactics from video sequences. It is based on players' trajectory during the play. They manually segmented the video into meaningful sub-sequences, represented the individual trajectories numerical strings, and used k-Nearest Neighbor~(k-NN) to classify and analyze player movement tactics. Ghosh~\emph{et al}.~\cite{ghosh2018smarttennistv} temporally segmented the rallies in the video using scores shown on the TV broadcasting screen and refined the recognized result using a tennis scoring system. They used a system to provide a simple interface for humans to retrieve and view a relevant video segment in the match. Ghosh~\emph{et al}.~\cite{ghosh2018towards} improved the temporal segmentation of the point using machine learning, both traditional and deep learning techniques, to classify a frame into a point or a non-point frame.

Our work has two major differences from the above prior works. Firstly, since we do not have access to player bounding boxes annotations during training, we propose a novel unsupervised temporal feature clustering method for high precision player retrieval. Secondly, our dataset shows a high degree of pose variation with limited color variation making our model liable to overfitting, which we address by using a Generative Adversarial Networks~(GAN) based image translation method.

\section{Table tennis video dataset}

Our video dataset consists of 72 videos of table tennis matches in total, with each video being approximately 1 hour long. Every video is for singles game The camera always points at the table with arbitrary elevation and azimuth of camera pose. We roughly define videos as \emph{straight} where the azimuth is between $-20 \deg$ to $20 \deg $ and \emph{oblique} if it is greater than $20 \deg$ by manual inspection. Among all the videos, $50$ were classified as \emph{straight} videos and other $22$ were oblique videos. We divided the total number of videos to 60 training, 6 testing and 6 validation samples. 6 testing and validation videos were divided into 3 \emph{straight} and 3 \emph{oblique} videos. Some screen-shots from the videos in our datasets are shown in Figure~\ref{fig:compare}~(c).

In table tennis, sports analysts are mostly interested in detecting rallies, which is defined as the duration when the ball is in play on the table. Our problem setting involves taking a raw input video as input and performing a temporal segmentation of those rally scenes in real-time. During training, we assume that annotations in the form of start and end time of rallies are only provided and no annotated bounding box information showing the player locations on the video are provided.

\section{Proposed Method}


\subsection{Unsupervised player retrieval}
\label{sec:retrieve}

In the presence of annotated player bounding boxes, prior method~\cite{ghosh2018towards} trains an object detector specifically for player detection. In our application, we do not have access to annotated bounding boxes~(BB) during training. Our application demands that we do not use any form of user interaction to annotate player BBs and hence we opt for a completely unsupervised approach for player retrieval. We use a pre-trained object detector to detect all BBs corresponding to ``person" class from each frame. Next, we use a heuristic score based ranking method to obtain the top two scoring BBs as the players. However, such a heuristics based score is weak and produce low precision rates. We propose a feature clustering-based approach to boost the performance of the heuristic score by identifying two dominant clusters in detected ROIs across multiple temporally separated frames, where we refer the cropped image inside each BB as Region Of Interest~(ROI).

\subsubsection{Table detection}
\label{sec:table}
Players usually occupy a position close to the table during a game and hence it is important to extract the location of the table as our first step. We find that the table is located close to the image center for most straight videos. However, this is not always the case for oblique videos, thus requiring a general table detection method for finding the distance between players and the table center. We used a MASK R-CNN~\cite{he2017mask} network pre-trained on MS-COCO~\cite{lin2014microsoft} dataset to obtain the mask corresponding to the table by detecting the labels of ``bench" class. Table centers $\bm{c} = (c_x, c_y)$ for each video are subsequently computed by weighted mean as, $c_x= \frac{\sum_i x_i m_i}{\sum_i m_i}$ and $c_y= \frac{\sum_i y_i m_i}{\sum_i m_i}$, where $m_i$ is the mask value (which is 1 for occupancy and 0 else) and  $x_i$, $y_i$ are absicca and ordinate of the $i^{th}$ pixel on the image.

\subsubsection{Heuristic score for player retrieval}
\label{sec:heuristic}
\begin{figure}[tb]
	\centering
	\includegraphics[width=0.45\textwidth]{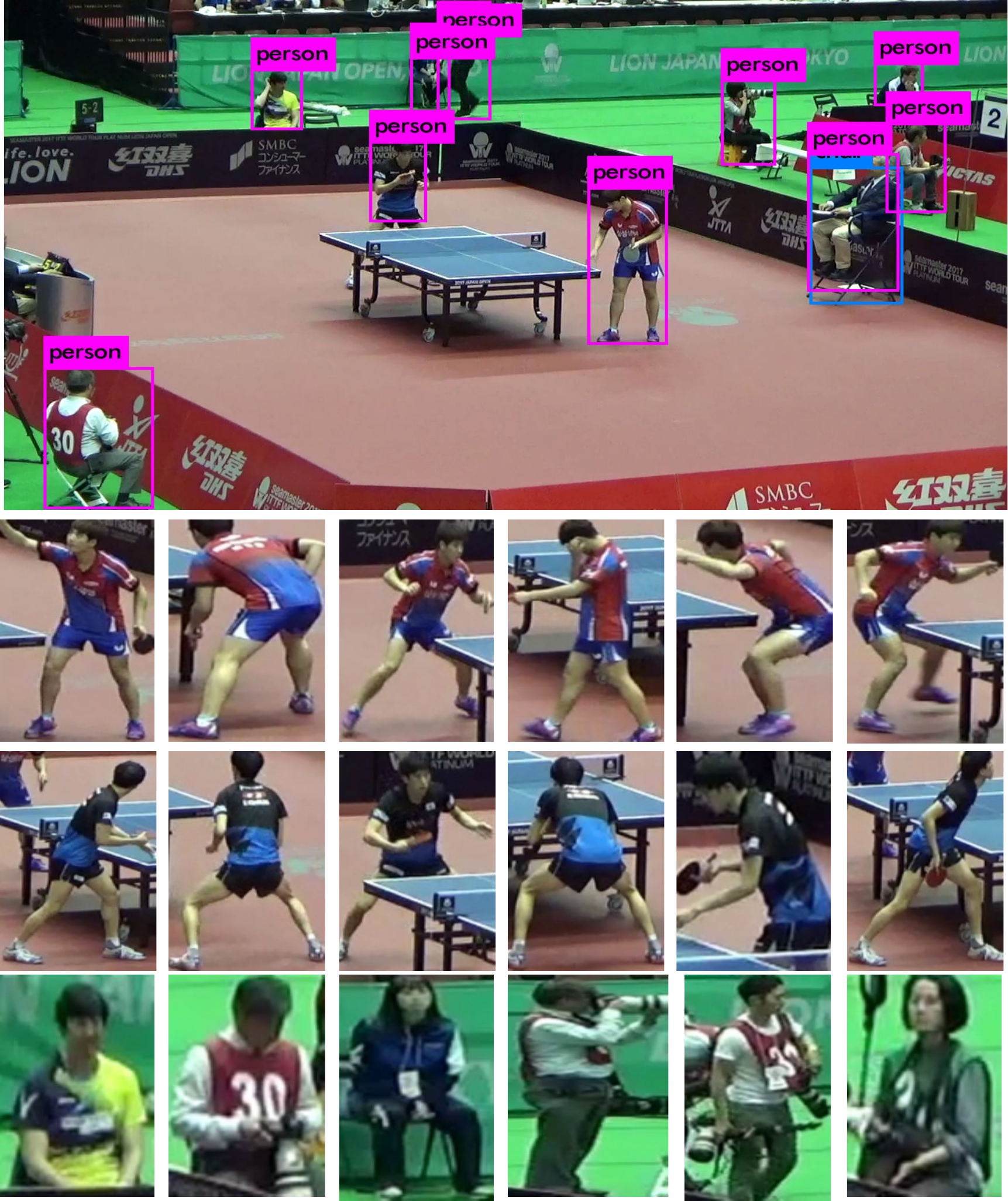}
	\caption{Illustration of unsupervised player retrieval from video frames. Top: Showing multiple ``Person" bounding box from pretrained object detector. Bottom: Showing samples obtained using temporal aggregation and clustering. We succesfully cluster players as two dominant clusters (top 2 rows) with highest scores and outliers as the third cluster~(bottom row).}
	\label{fig:clustering}
\end{figure}

We use a pre-trained object detector network trained on MSCOCO~\cite{lin2014microsoft} dataset and use it for person BB detection in each frame. As seen in Figure~\ref{fig:clustering}, multiple person BBs are detected by the object detector and our objective is to only retain the two BBs corresponding to the players. 

Let $I^f_k$ be the $k^{th}$ ROI for the $f^{th}$ frame. We assign a score to each ROI based on some heuristics. Firstly, we found that in most cases, the probability that detected ROI is ``Person"~($p(h|I^f_k)$) is higher for the player BBs compared to the other peripheral detections. Secondly, players usually occupy the playing area in close proximity to the table center~$(c_x, c_y)$ and hence player BB distance from the table center should be less compared to other BBs. We use a weighted combination of these two factors to compute a score for the $k^{th}$ bounding box in each frame, given as 

\vspace*{-0.25cm}

\begin{equation}
	S_\mathcal{H}^f(k) = \alpha  {p(h|I^f_k)} + \beta  \big(1 - \frac{||\bm{x}_k - \bm{c}||_2}{||\bm{c}||_2} \big)
\end{equation}

\vspace*{-0.25cm}

\noindent where $\alpha$, $\beta$ are positive coefficeints and $\bm{x}$ is the center of $k_{th}$ BB. Top two bounding boxes having the highest score in each video frame, are chosen for player retrieval. However, this heuristics based method produces considerable amount of false positive detections reducing the precision of player retrieval. Next, we describe the clustering based score boosting method for improving the precision of detection.

\subsubsection{Score boosting by temporal feature aggregation} 
\label{sec:cluster}
The heuristic scoring method described in the previous section is a weak metric for ranking detected BBs for players extraction resulting in numerous outlier retrievals as shown in Figure~\ref{fig:clustering}. To mitigate this problem, we exploit temporal information in retrieved BB distribution across frames and perform a clustering-based outlier rejection steps to boost the performance above heuristic score. The idea is to detect clusters on all detected BBs across the temporal axis of the video and extract which two dominant clusters have the highest cumulative score distribution. Using this information, we update the heuristic score to include a similarity measure to the cluster centers which is then used to rerank the detected BBs to remove major outliers. 

We uniformly sample $M$ frames for each video assuming that in most frames the two players will be present close to the table (while in the game). For each frame, we crop all the ``person" ROIs and rank them using the heuristic score based method retaining the top two candidates. These ROIs are pooled together across frames to get a total of $2M$ frames and stored in a buffer, $I_B = \{I_0^f, I_1^f\}_{f=1 \;\text{to} \;M}$ . The goal now is to perform clustering and estimate the two cluster centers corresponding to the two players.

From the cropped ROIs we extract person reidentification features. Since such features are invariant for varying human poses, features corresponding to the players are likely to form clusters for each person. We use a part model based person reidentification feature extractor~\cite{sun2018beyond} which exhibits state-of-the-art performance.  Similarly features corresponding to outliers also form one or many small clusters. We represent the person reidentification transformation as $g_{person}$ and the features are computed as $X_R = g_{person}(I_B^f)|_{f=1\; \text{to}\; 200}$. Thus we represent the features space distribution as a mixture of Gaussians, where the maximum number of component Gaussians, $N_{max}$ is unknown,

\vspace*{-0.2cm}

\begin{equation}
p(X_R) = \underbrace{\alpha_1 \mathcal{N}(\bm{\mu}_1, \bm{\sigma}_1)}_{\text{Player 1 cluster}} + 
\underbrace{\alpha_2 \mathcal{N}(\bm{\mu}_2, \bm{\sigma}_2)}_{\text{Player 2 cluster}} + 
\underbrace{\sum_{k=3}^{N_{max}} \alpha_k \mathcal{N}(\bm{\mu}_k, \bm{\sigma}_k)}_{\text{Outlier cluster}}
\end{equation}

The objective is to discover the player cluster centers $\{\bm{\mu}_1, \bm{\mu}_2\}$, which can be used to boost the weak heuristic scores for better player retrieval. We initially cluster the feature space using Expectation Maximization, with the number of clusters set to 3. It is assumed that two clusters with the highest cumulative heuristic score will be the player clusters. However, we found some cases where the two player ROIs are very similar (due to similar jerseys, lighting conditions, etc.) and they are clustered together as the first dominant cluster with outlier ROIs detected as the second cluster. To alleviate this situation, we use a fitness measure that ensures that two dominant clusters have almost equal score distribution. Thus, if one cluster contains player ROIs, which have a high score on average, and the second cluster has predominantly outliers having low scores, the fitness measure will be much smaller than the healthy value of 1.  In such a case, we again perform clustering eliminating the feature points in the lowest scoring cluster. This method is described in algorithm \ref{alg:temporal}.

\begin{algorithm*}[h]
	\caption{Temporal feature aggregation for heuristic score boosting}
	\label{alg:temporal}
	\begin{algorithmic}[1]
		\REQUIRE Video input, low and high thresholds $\{t_l, t_h\}$ for fitness measure 
		\STATE Extract $M$ video frames, $\{I^f\}_{j=f\;to\;M+f}$
		\FOR{$i$ from $0$ to $M$}
		\STATE    - Extract top two BB candidates, $\{I_0^f, I_1^f\}$ using heuristic score $S_\mathcal{H}$ and \\store the corresponding ROIs and score to buffer, $I_B \leftarrow \{I_0^f, I_1^f\}$ and $S_B \leftarrow \{S_{\mathcal{H}(0)}^i, S_{\mathcal{H}(1)}^i\}$  \\
		\ENDFOR
		
		\STATE Extract person re-identification features for the ROIs, $X_R^f = g_{person}(I_B^f)|_{f=1\; to\; 2M}$
		
		\STATE Initialize fitness score, $S_{fitness} \leftarrow 0$, number of cluster, $N_{cluster} \leftarrow 3$
		
		\WHILE{$S_{fitness}  \le t_l$ or $S_{fitness}  \ge t_h$}
		\STATE - Compute Expectation Maximization on person re-identification features $X_R$, to find cluster centers, $\{\bm{\mu}_i\}_{i=1\;to\;N_{cluster}}$
		\STATE - Compute mean score of all the points that belong to cluster $j$ and sort them in descending order to get $\{S_j^{sorted}\}_{j=1\;to\;N_{cluster}}$ and sorted cluster center as $\{\bm{\mu}_i^{sorted}\}_{i=1\;to\;N_{cluster}}$
		
		\STATE - Compute fitness score, $S_{fitness} = \frac{S_1^{sorted}}{S_0^{sorted}}$, set $N_{cluster} \leftarrow 2$. Eliminate the feature points for the lowest scoring cluster.
		
		\ENDWHILE
		\STATE Compute boosted score for the $k^{th}$ bounding box for $f^{th}$  image frame as 
		$S_{boosted}^f(k) = S_{\mathcal{H}}^f(k) - \kappa \min(||\bm{x}_R^f(k) - \bm{\mu}_0^{sorted}||, ||\bm{x}_R^f(k) - \bm{\mu}_1^{sorted}||)$, where $\bm{x}_R^f(k) = g_{person}(I_f^j)$ is the person reid feature for $k^{th}$ ROI.
	\end{algorithmic}
\end{algorithm*}

The above algorithm extracts the player clusters having the highest cumulative score. In our experiments, we found that K-Means and Expectation Maximization can be used interchangeably to yield similar results. The old heuristic score is updated by a distance measure from the dominant player cluster centers, $\kappa \min(||\bm{x}_R^f(k) - \bm{\mu}_0^{sorted}||, ||\bm{x}_R^f(k) - \bm{\mu}_1^{sorted}||)$, thus boosting the performance by using additional appearance based similarity measure. Usually, the value of $\kappa$ is kept high to give more weight to the component of the score computing distance from cluster centers. The temporal feature clustering is performed as a pre-processing step and the boosted score from the above algorithm is used to rank the detected BBs in each frame for player retrieval. This process is illustrated in Figure \ref{fig:clustering}.

\begin{figure}[b]
	\centering
	\includegraphics[width=0.46\textwidth]{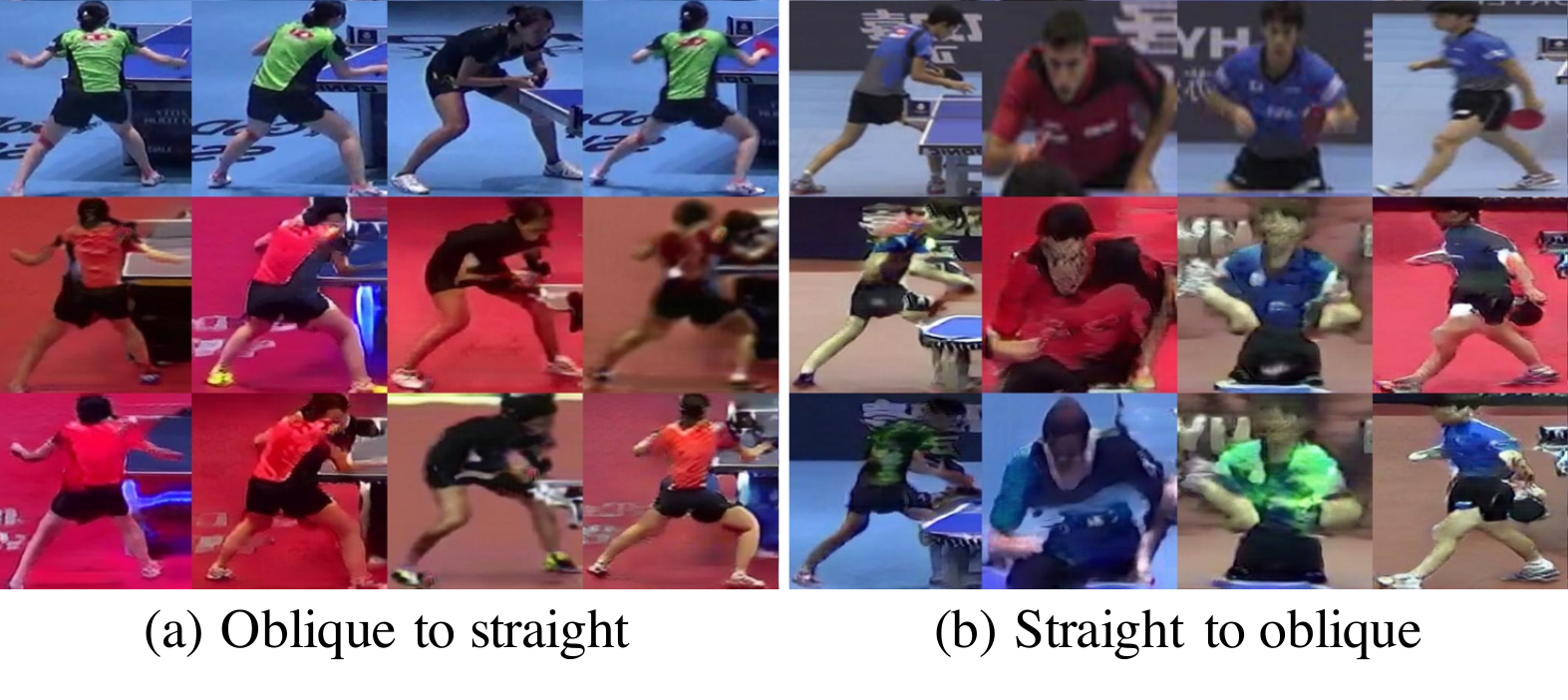}
	\caption{Sample generated images using GAN-based multi-modal image translation model. Top row shows true training samples and last two rows shows generated images with same content information but randomly sampled style vectors. }
	\label{fig:munit}
\end{figure}

\subsection{Training samples}
\label{sec:training}
We extract training samples by extracting small contiguous sub-video windows around the annotated start and endpoints for each rally in the video. 
We use windowing around the start point for positive samples and those around the endpoint for negative samples. This forces the model to produce high detection probability at the start of the rally and reduce the probability score at the endpoints. Additionally, we also use randomly sampled points in the ``non-rally" regions to mine negative examples.  Since the variation in the negative samples is much larger (players walking, resting, or any other action) compared to the positive samples (which mostly constitutes of a serve, shots played followed by the end of rally), 
the number of negative samples in our dataset is five times the number of positive samples. We used a window of size $\pm 4$ seconds around the annotated points for sample extraction.

Our goal is to learn a binary classifier that detects whether a sample is in a rally or not. During inference, this model is used in a sliding window fashion to detect the temporal location of rallies in testing videos. We classify the extracted features using a sigmoid activated LSTM~\cite{lstm} based video activity recognition network inspired from \cite{ullah2018action}. We found that bidirectional LSTM did not give us much improvement in performance and hence we used uni-directional LSTM in our application. 


\subsection{Multi-modal image translation for improved generalization}
\label{sec:augment}

The number of obliquely angled videos in our training set is limited to 16 compared to 44 straight capture videos. Although there is a high variation of camera pose in these few oblique videos, there is limited variation in the color distribution which poses a risk of over-fitting to the training set. 

To alleviate this problem, we present a data augmentation scheme that introduces variability in image appearance across training samples by artificially generating images using GANs. Our objective is to only introduce variations in appearance while keeping the content (the players pose and action) of the samples unchanged. Multi-modal Unsupervised Image Translation (MUNIT) model~\cite{huang2018multimodal} is used for data generation. We arrange the positive training samples for straight and oblique videos into two domains and image translation is performed between them. 

\begin{figure}[tb]
	\centering
	\begin{subfigure}[tb]{.45\textwidth}
		\centering
		\normalsize
		\begin{tabular*}{\textwidth}{c @{\extracolsep{\fill}} l|rrr|}
			&& \multicolumn{2}{c}{Average Precision (AP)}\\
			\hline
			&& Heuristic & Boosted\\
			\hline
			Oblique && {0.78} & \textbf{0.86} \\
			Straight && 0.95 & \textbf{0.98} \\
			Combined && 0.86 & \textbf{0.92}\\
			\hline
		\end{tabular*}
		\caption{ Player retrieval Average Precision~(AP) for heuristic and boosted scores by temporal aggregation.}
	\end{subfigure} 
	\vspace*{2mm}
	
	\begin{subfigure}[]{.35\textwidth}
		\label{fig:pr}
		\centering
		\includegraphics[width=\textwidth]{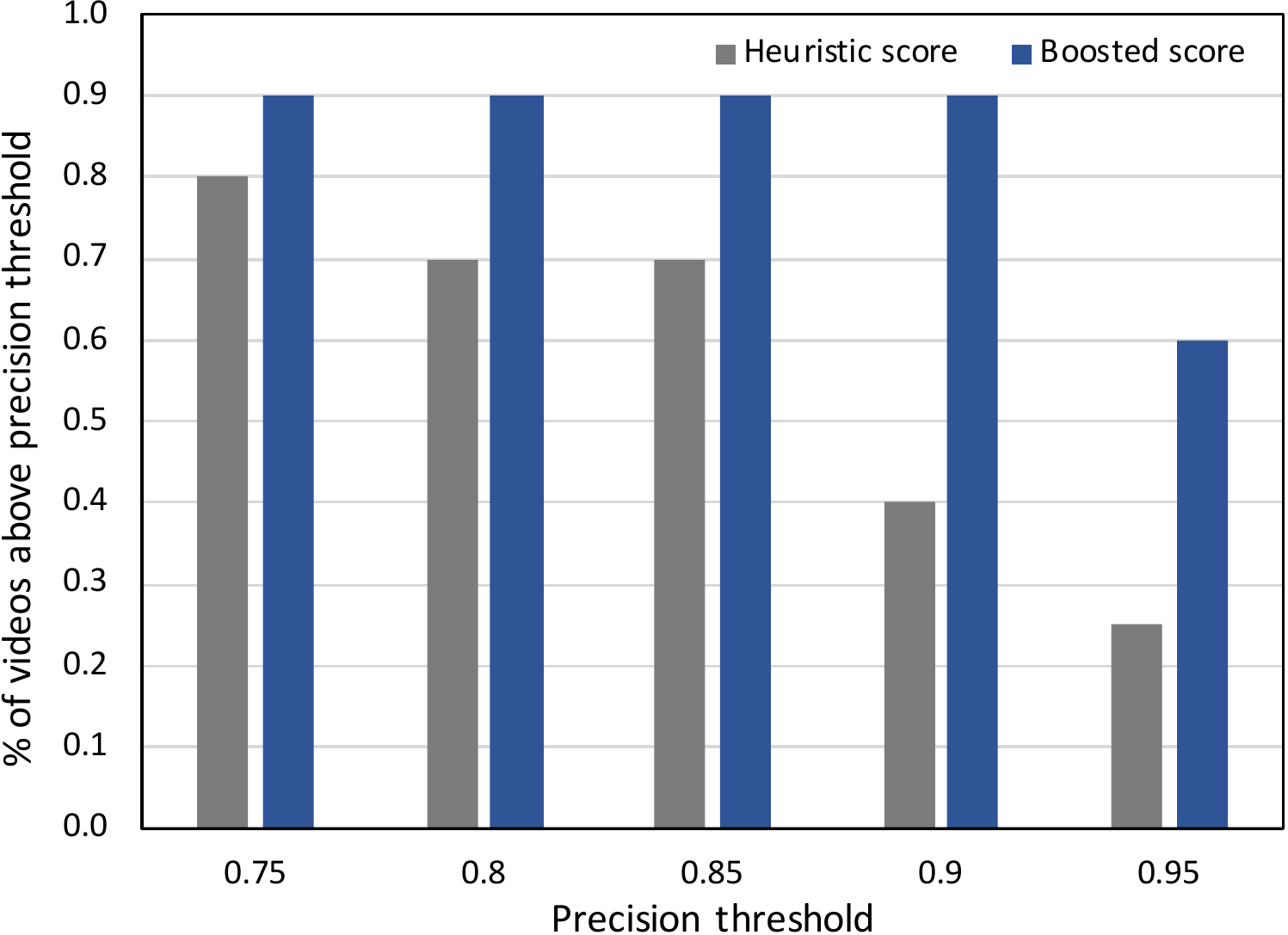}
		\caption{\small Percentage of videos having precision greater than threshold on $x$-axis. More \% at higher threshold is better.}
	\end{subfigure}%
	
	\caption{\small \textbf{Evaluation of unsupervised player retrieval.}}
	\label{art}
\end{figure}

Given two domains $\{1,2\}$, the main idea is to factorize the latent representations for an image into a content code $c_{i}$ and a style code $s_{i}$, using the encoder networks as $(c_{i}, s_{i}) = (E_{i}^{c}(x_{i}), E_{i}^{s}(x_{i}))$, where $E_{i}^{c}(.)$ and $E_{i}^{s}(.)$ are the content and style encoder respectively for the $i^{th}$ domain. With $G_i$ as the decoder for the $i^{th}$ domain and $q(s_1), q(s_2)$ are prior distribution of style code for respective domains, training is performed by minimizing three kinds of loss: image reconstruction loss, latent space reconstruction loss and adversarial loss for cross-domain distribution matching. During inference, image-to-image translation is performed by using content encoder of one domain, and a randomly drawn style code from the other domain, as $x_{1\rightarrow 2} = G_{2}\big(E_{1}^{c}(x_1), s_{2}\big)\big|_{s_2 \sim q(s_2)}$ and vice versa. For more details on the image translation model, we refer the reader to the original paper by Huang et al.~\cite{huang2018multimodal}. Figure \ref{fig:munit} shows generated samples for two randomly drawn style codes for both directions of straight to oblique videos and vice versa.

To perform data augmentation, we generate one copy of synthetic data for all positive training samples ensuring that the style code remains the same for all image frames in each sample. As a result of this data augmentation, we get a double number of positive samples improving the variability in appearance features in the learning data. It is to be noted that we can use an arbitrarily large number of style codes to improve appearance variability even more, however that would require more memory resources.

\section{Experimental Results}
\label{sec:exp}
In this section, we explain the baseline methods used and discuss the quantitative and qualitative results. We perform several experiments to seek answers to the following questions,
\begin{itemize}
	\item [Q1:] Does the temporal feature aggregation based score boosting improve the precision of player retrieval?
	\item [Q2:] Are player-level features better for rally detection than frame-level global features?
	\item [Q3:] Does data augmentation using a multi-modal image translation model improve F1 score by reducing over-fitting?
	\item [Q4:] What is the precision and recall of the trained model in terms of temporal segmentation accuracy?
\end{itemize}

\begin{table*}[t]
	\centering
	\normalsize    
	\caption{Sample-wise evaluation on test set. First two rows use frame-level~(baseline), last three rows use player-level~(proposed) features. ``+" means addition of train setting to previous row. Proposed models give better F1 score and AUROC compared to baseline.~(``Pr": precision, ``R": Recall. ``st" and ``obl": training with straight and oblique videos).}
	\begin{tabular*}{\textwidth}{c @{\extracolsep{\fill}} l|r|rrr|rrr|rrr|}
		\hline
		&& \multicolumn{1}{c|}{\textbf{AUROC}} & \multicolumn{3}{c|}{\textbf{Oblique}} & \multicolumn{3}{c|}{\textbf{Straight}} & \multicolumn{3}{c|}{\textbf{Combined}}\\ 
		&&  & Pr & R & F1 & Pr & R & F1 & Pr & R & F1\\ \hline
		\textbf{Frame-level (st)} && {0.950} & {0.51} & {0.79} & {0.590} & {0.81}  & {0.92} & \textbf{{0.86}} & {0.62} & {0.85} & {0.71}\\
		\textbf{Frame-level (+ obl)} && {0.978} & {0.56} & {0.92} & {0.70} & {0.70}  & \textbf{{0.99}} & {0.82} & {0.62} & \textbf{{0.95}} & {0.75}\\
		\textbf{Proposed (player-level, st)} && {0.984} & \textbf{{0.93}} & {0.76} & {0.83} & {0.80}  & {0.92} & {0.85} & {0.85} & {0.84} & {0.85}\\
		\textbf{Proposed (+ obl)} && {0.989} & {0.91} & {0.9} & \textbf{{0.91}} & \textbf{0.82}  & {0.85} & {0.83} & \textbf{{0.86}} & {0.88} & {0.87}\\
		\textbf{Proposed (+ aug)} && \textbf{0.993} & {0.87} & \textbf{{0.95}} & \textbf{{0.91}} & {0.80}  & {0.89} & {0.84} & {0.84} & {0.92} & {\textbf{0.88}}\\ \cline{1-12} 
	\end{tabular*}
	
	\label{table:sample}
\end{table*}

\subsection{Experimental settings}
\label{sec:settings}
We set up baseline methods for evaluating the performance of our method. To examine the necessity of using unsupervised player retrieval, we consider two cases. In the first case, where we do not use player BB information and train our detector on global frame-level features and in the second case we use our proposed player-level features to train the detector as described in Section~\ref{sec:training}. We perform training on the above models in both cases of straight camera captured videos and on the combination of straight and obliquely angled videos. Additionally, we also train with multimodal image translation based data augmentation.

We used VGG19~\cite{simonyan2014very} networks, pre-trained on the ImageNet dataset \cite{deng2009imagenet} for feature extraction. Additional, we also tried ResNet~\cite{he2016deep} but VGG19 produced better results. Video frames were resized to $224 \times 224 \times 3$ RGB images, for extracting global frame-wise features in case of the baseline methods. For player-level features, we used the YOLO object detector \cite{redmon2016you} pre-trained on MSCOCO dataset \cite{lin2014microsoft}. 
For the LSTM based classifier, we use $2$ hidden layers, each with $256$ and $64$ $\tanh$-activated units respectively, followed by a fully connected layer of one unit output which is sigmoid activated. We used a frameskip of $5$ frames. Therefore, each data point in our training is a tensor of size $41 \times 224 \times 224 \times 3$ which after computing VGG features is transformed to size $41 \times 1024$ dimensional tensor.

\subsection{Evaluation of unsupervised player retrieval}

To evaluate the precision of our proposed unsupervised player retrieval, we randomly select 10 oblique and 10 straight videos. For each video, we uniformly sample $M=100$ frames and perform extraction of two players for each such frame using heuristic score method and clustering based boosted score method. We manually compute the number of false positive (FP) detections (images that are not players, eg. coaches, audience, etc.) from the cropped image ROI. Precision is defined as 
$Pr = ({\text{N}_{\text{crops}} - \text{FP}})/{{\text{N}_{\text{crops}}}}$, where $\text{N}_{\text{crops}} = 200$ in this case. We report the average precision value for oblique and straight video cases in Table \ref{art}(a). 

The proposed score boosting method has a high player retrieval rate compared to the heuristic score for every setting. Particularly for the oblique case, the score-boosting method significantly improves the precision of player detection from 0.78 to 0.86, showing that heuristic score based method produces numerous outlier detections that are successfully removed by the proposed method. Additionally, we also plot the percentage of videos (both oblique and straight) having precision greater than thresholds of 0.75 to 0.95 in intervals of 0.5 as shown in Figure \ref{art}~(b). The proposed score boosting method is shown to have a greater percentage of videos having high precision making it suitable for our application of unsupervised player extraction.

\subsection{Evaluation of rally detection} 
In our experiments, we use two kinds of evaluations to compare the efficiency of our method to the baseline. We use precision, recall, and F1-score as the metric of comparison in both cases. 

\subsubsection{Sample-wise evaluation} 
\begin{figure*}[t]
	\centering
	\begin{subfigure}[]{.3\textwidth}
		\label{fig:pr}
		\centering
		\includegraphics[width=\textwidth]{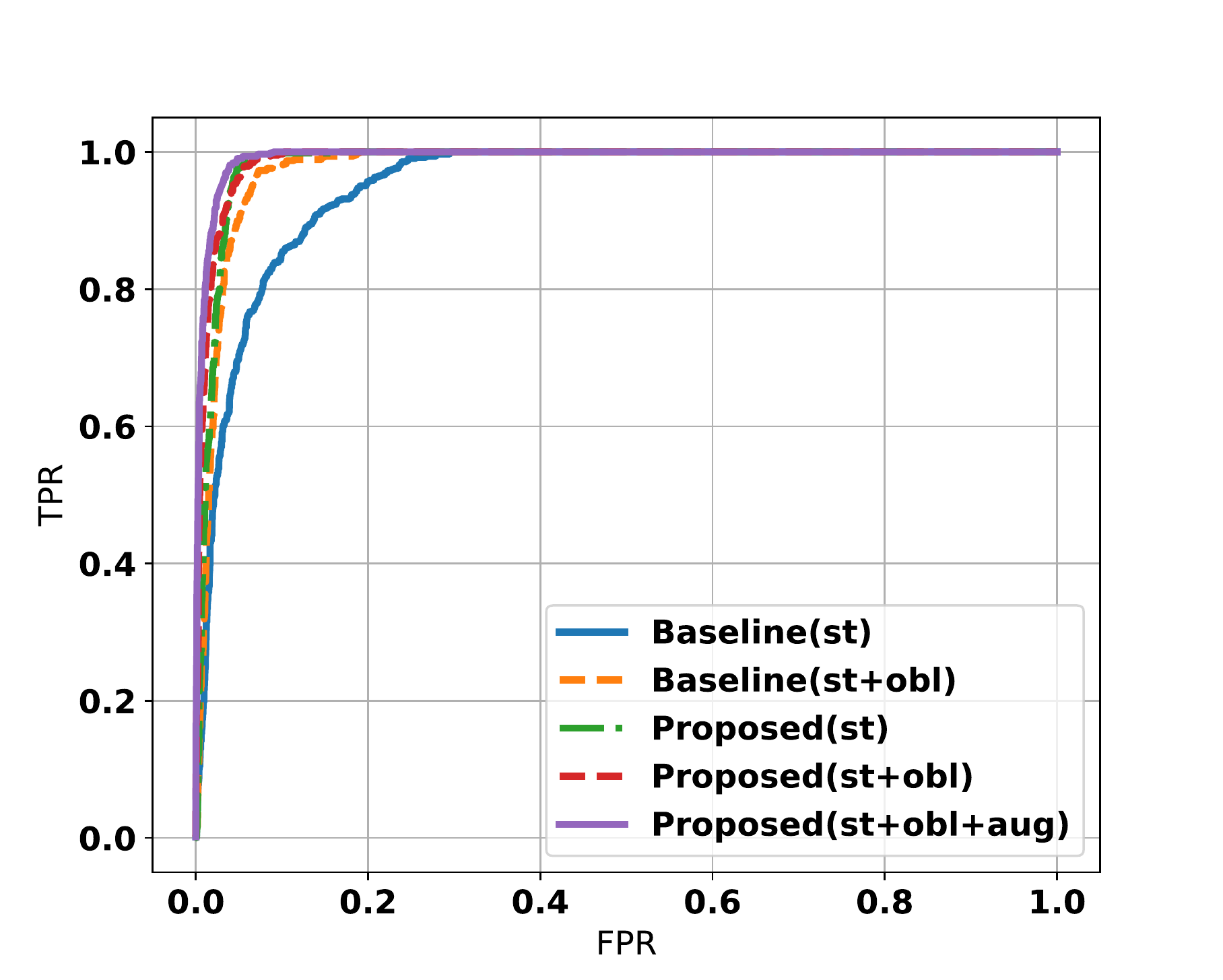}
		\caption{\small ROC-Curve}
	\end{subfigure}%
	\begin{subfigure}[]{.3\textwidth}
		\label{fig:pr}
		\centering
		\includegraphics[width=\textwidth]{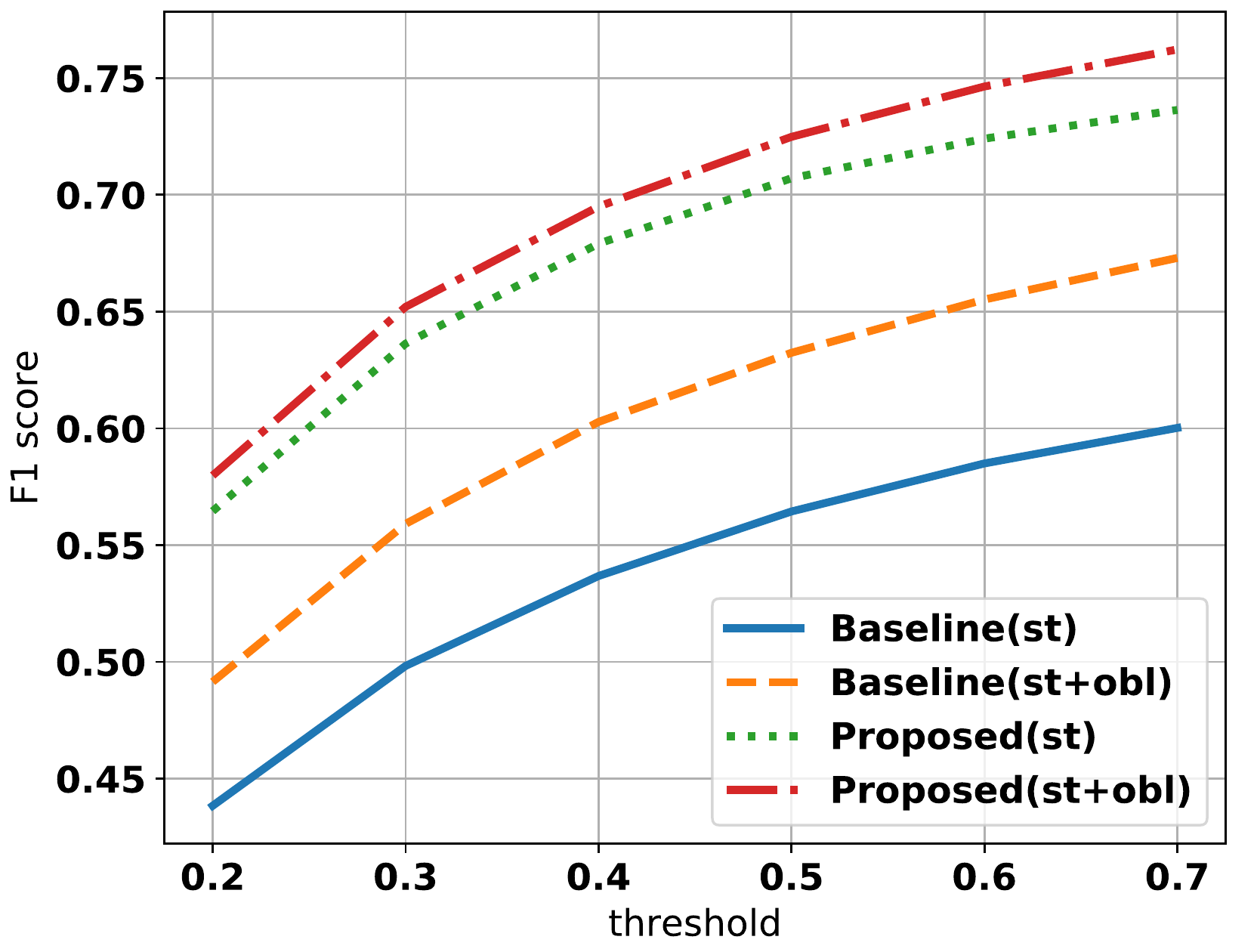}
		\caption{\small Termporal Evaluation (Serve)}
	\end{subfigure}%
	\begin{subfigure}[]{.3\textwidth}
		\label{fig:pr}
		\centering
		\includegraphics[width=\textwidth]{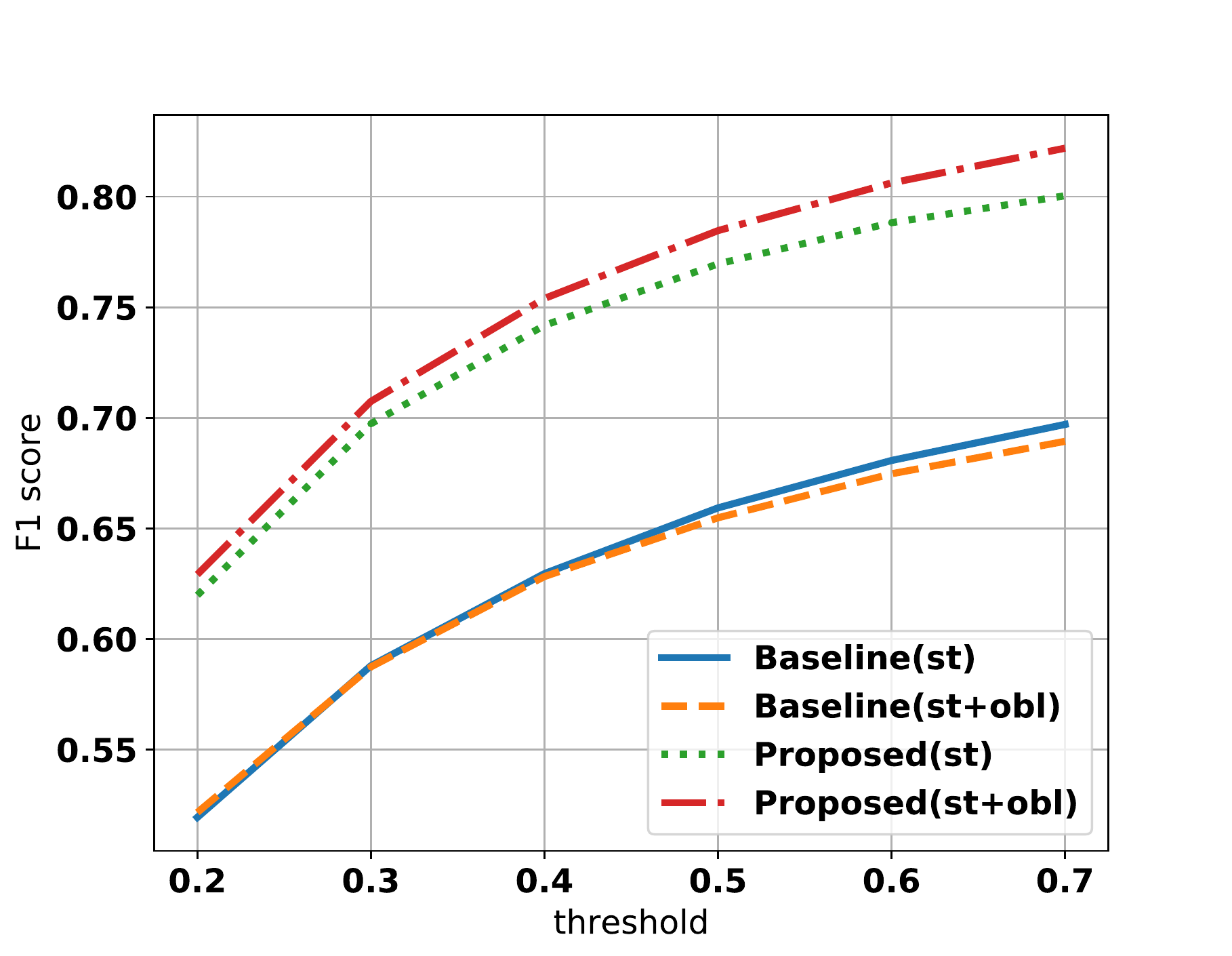}
		\caption{\small Termporal Evaluation (End of rally)}
	\end{subfigure}%
	
	\caption{\small \textbf{Evaluation of proposed method, (a) Our method has best AUROC score compared to baseline methods. Plot of F1 score at various thresholds for (b) Serve point detection (c) End-of-rally detection.}}
	\label{fig:temporal}
\end{figure*}

Here, we use our learned classifier to predict the class labels on the test samples.  
Performance on test set is shown in Table \ref{table:sample} and Figure~\ref{fig:temporal}(a) shows the ROC plot. Our proposed player-level feature-based method produces an F1 score of 0.87 compared to 0.75 for the model trained on frame-level features and has the best AUROC when compared to baseline. Thus, player-level features generalize better compared to global frame-level features. Ablation studies show that adding oblique videos in the training set, improves performance on oblique test cases, while performance on straight cases decreases. The proposed data augmentation method is also shown to improve F1 score. A possible reason for the limited improvement may be attributed to the lack the color variation between original training and test sets in our particular dataset. 



\subsubsection{Temporal segmentation evaluation}
This evaluation is tailored to find how well the detector localizes rally detection in time. For that purpose, we first apply the learned rally detector in a sliding window fashion to the entire length of the input video and record the confidence of the rally detection at each timestep.  The sliding window method provides contiguous confidence regions in time depicting the duration of a single rally which are thresholded and the rising and falling edges are detected as serve points and the end-of-rally point. To evaluate the serve and end-of-rally detection accuracy, we consider a tolerance window around the human-annotated serve point and end-of-rally point of $\pm$ 3 seconds. Point detections that lie within such a tolerance window are considered as true positive detections. The other positively detected points are false positives. 

We use a  range of threshold values and plot the corresponding F1 scores in Figure~\ref{fig:temporal}~(a) and Figure~\ref{fig:temporal}~(b) for serve and end-of-rally detections respectively. Our proposed model, trained on both straight and oblique videos, consistently performs better than the baseline method for all thresholds thus showing its superiority in performance. Our proposed method outperforms the baseline models~(using frame-level features) due to robust local player-level features. The frame-level methods are inefficient for capturing robust image features and therefore are not invariant to viewpoint transformations. We believe that the baseline methods overfit to the dominant mode of the overall distribution(straight cam videos) and fail to generalize to the other less common distribution (oblique cam videos). Additionally, we found that a potential source for false-positive detections for all methods was ``let" events where the rally may prematurely end due to a ``foul". These events were not marked as rally scenes in ground truth annotations however since the players perform ``serve" action, these were detected as a rally by our detector. Please see the attached supplementary video for manual inspection of detection quality.

%
\subsubsection{Robustness to shaking camera and occlusion}

We synthetically simulate the effect of camera shaking by randomly cropping an area in the image from each frame. We also simulated occlusions by artificially adding an occluding box of size (0.4$H$, 0.2$H$) at random positions in all the video frames in the test set, where $H$ is the height of the video frame. These simulated perturbations are used to test our model's performance under highly noisy real-world scenarios. Qualitative analysis from attached supplementary videos shows that our method detects most rally scenes even in the presence of noise, although in the presence of occlusion noise, we find some false positive detections. We believe that our LSTM based classifier possibly learned to extract discriminative features only from a few key-frames of the input window. Thus, in spite of erroneous features from some frames, it can robustly pick up discerning features from other frames, thus maintaining the robustness of detection.  
%

\section{Conclusion}

We presented a robust method for event detection in unstructured sports videos. Firstly, we proposed a novel temporal feature aggregation method for boosting heuristic scores to obtain high precision player retrieval rates. Additionally, we perform data augmentation using multi-modal image translation to reduce bias in appearance-based features in the training set. Our method produces accurate unsupervised player extraction which in turn is used for precise temporal segmentation of rally scenes. Additionally, our method is robust to noise in the form of camera shaking and occlusion, which enables its applicability to videos captured from low-cost commercial cameras. Although we present an application in table tennis, our method can be applied to general racquet games like badminton, tennis etc. In the future, we want to extend this work to videos of doubles games with fine-grained action recognition for detecting various kinds of shots in an unstructured setting. Other extensions might include analyzing unstructured videos of non-racquet games like soccer, rugby, etc.

\bibliographystyle{IEEEtran}
\bibliography{sports}

\begin{thebibliography}{10}
\providecommand{\url}[1]{#1}
\csname url@samestyle\endcsname
\providecommand{\newblock}{\relax}
\providecommand{\bibinfo}[2]{#2}
\providecommand{\BIBentrySTDinterwordspacing}{\spaceskip=0pt\relax}
\providecommand{\BIBentryALTinterwordstretchfactor}{4}
\providecommand{\BIBentryALTinterwordspacing}{\spaceskip=\fontdimen2\font plus
\BIBentryALTinterwordstretchfactor\fontdimen3\font minus
  \fontdimen4\font\relax}
\providecommand{\BIBforeignlanguage}[2]{{%
\expandafter\ifx\csname l@#1\endcsname\relax
\typeout{** WARNING: IEEEtran.bst: No hyphenation pattern has been}%
\typeout{** loaded for the language `#1'. Using the pattern for}%
\typeout{** the default language instead.}%
\else
\language=\csname l@#1\endcsname
\fi
#2}}
\providecommand{\BIBdecl}{\relax}
\BIBdecl

\bibitem{vgg}
K.~Simonyan and A.~Zisserman, ``Very deep convolutional networks for
  large-scale image recognition,'' \emph{CoRR}, vol. abs/1409.1556, 2014.

\bibitem{resnet}
\BIBentryALTinterwordspacing
K.~He, X.~Zhang, S.~Ren, and J.~Sun, ``Deep residual learning for image
  recognition,'' in \emph{2016 {IEEE} Conference on Computer Vision and Pattern
  Recognition, {CVPR} 2016, Las Vegas, NV, USA, June 27-30, 2016}, 2016, pp.
  770--778. [Online]. Available: \url{https://doi.org/10.1109/CVPR.2016.90}
\BIBentrySTDinterwordspacing

\bibitem{densenet}
G.~Huang, Z.~Liu, L.~van~der Maaten, and K.~Q. Weinberger, ``Densely connected
  convolutional networks,'' in \emph{Proceedings of the IEEE Conference on
  Computer Vision and Pattern Recognition}, 2017.

\bibitem{redmon2016you}
J.~Redmon, S.~Divvala, R.~Girshick, and A.~Farhadi, ``You only look once:
  Unified, real-time object detection,'' in \emph{Proceedings of the IEEE
  conference on computer vision and pattern recognition}, 2016.

\bibitem{fasterrcnn}
\BIBentryALTinterwordspacing
S.~Ren, K.~He, R.~Girshick, and J.~Sun, ``Faster r-cnn: Towards real-time
  object detection with region proposal networks,'' in \emph{Advances in Neural
  Information Processing Systems 28}, C.~Cortes, N.~D. Lawrence, D.~D. Lee,
  M.~Sugiyama, and R.~Garnett, Eds.\hskip 1em plus 0.5em minus 0.4em\relax
  Curran Associates, Inc., 2015, pp. 91--99. [Online]. Available:
  \url{http://papers.nips.cc/paper/5638-faster-r-cnn-towards-real-time-object-detection-with-region-proposal-networks.pdf}
\BIBentrySTDinterwordspacing

\bibitem{ssd}
\BIBentryALTinterwordspacing
W.~Liu, D.~Anguelov, D.~Erhan, C.~Szegedy, S.~Reed, C.-Y. Fu, and A.~C. Berg,
  ``Ssd: Single shot multibox detector,'' 2015, cite arxiv:1512.02325Comment:
  ECCV 2016. [Online]. Available: \url{http://arxiv.org/abs/1512.02325}
\BIBentrySTDinterwordspacing

\bibitem{simonyan2014two}
K.~Simonyan and A.~Zisserman, ``Two-stream convolutional networks for action
  recognition in videos,'' in \emph{Advances in neural information processing
  systems}, 2014, pp. 568--576.

\bibitem{tran2015learning}
D.~Tran, L.~Bourdev, R.~Fergus, L.~Torresani, and M.~Paluri, ``Learning
  spatiotemporal features with 3d convolutional networks,'' in
  \emph{Proceedings of the IEEE international conference on computer vision},
  2015.

\bibitem{yoshikawa2010automated}
F.~Yoshikawa, T.~Kobayashi, K.~Watanabe, and N.~Otsu, ``Automated service scene
  detection for badminton game analysis using chlac and mra,'' \emph{World
  Academy of Science, Engineering and Technology}, 2010.

\bibitem{yan2014automatic}
F.~Yan, J.~Kittler, D.~Windridge, W.~Christmas, K.~Mikolajczyk, S.~Cox, and
  Q.~Huang, ``Automatic annotation of tennis games: An integration of audio,
  vision, and learning,'' \emph{Image and Vision Computing}, vol.~32, no.~11,
  pp. 896--903, 2014.

\bibitem{ghosh2018towards}
A.~Ghosh, S.~Singh, and C.~Jawahar, ``Towards structured analysis of broadcast
  badminton videos,'' in \emph{2018 IEEE Winter Conference on Applications of
  Computer Vision (WACV)}.\hskip 1em plus 0.5em minus 0.4em\relax IEEE, 2018,
  pp. 296--304.

\bibitem{huang2018multimodal}
X.~Huang, M.-Y. Liu, S.~Belongie, and J.~Kautz, ``Multimodal unsupervised
  image-to-image translation,'' in \emph{Proceedings of the European Conference
  on Computer Vision (ECCV)}, 2018, pp. 172--189.

\bibitem{lea2016segmental}
C.~Lea, A.~Reiter, R.~Vidal, and G.~D. Hager, ``Segmental spatiotemporal cnns
  for fine-grained action segmentation,'' in \emph{European Conference on
  Computer Vision}.\hskip 1em plus 0.5em minus 0.4em\relax Springer, 2016, pp.
  36--52.

\bibitem{Ramanathan_CVPR2016}
V.~Ramanathan, J.~Huang, S.~Abu-El-Haija, A.~Gorban, K.~Murphy, and L.~Fei-Fei,
  ``Detecting events and key actors in multi-person videos,'' pp. 3043--3053,
  06 2016.

\bibitem{SINGH201745}
\BIBentryALTinterwordspacing
S.~Singh, C.~Arora, and C.~Jawahar, ``Trajectory aligned features for first
  person action recognition,'' \emph{Pattern Recognition}, vol.~62, pp. 45 --
  55, 2017. [Online]. Available:
  \url{http://www.sciencedirect.com/science/article/pii/S0031320316301893}
\BIBentrySTDinterwordspacing

\bibitem{ghanem2018activitynet}
B.~Ghanem, J.~C. Niebles, C.~Snoek, F.~C. Heilbron, H.~Alwassel, V.~Escorcia,
  R.~Khrisna, S.~Buch, and C.~D. Dao, ``The activitynet large-scale activity
  recognition challenge 2018 summary,'' \emph{arXiv preprint arXiv:1808.03766},
  2018.

\bibitem{Reno_2018_CVPR_Workshops}
V.~Reno, N.~Mosca, R.~Marani, M.~Nitti, T.~D'Orazio, and E.~Stella,
  ``Convolutional neural networks based ball detection in tennis games,'' in
  \emph{The IEEE Conference on Computer Vision and Pattern Recognition (CVPR)
  Workshops}, June 2018.

\bibitem{Tom_2014}
T.~Dierickx, ``Badminton game analysis from video sequences,'' Master's thesis,
  Universiteit Gent. Faculteit Ingenieurswetenschappen en Architectuur., 2014,
  [Online; accessed 01-September-2018].

\bibitem{Vuckovic_2014}
\BIBentryALTinterwordspacing
G.~Vučković, N.~James, M.~Hughes, S.~Murray, Z.~Milanović, J.~Perš, and
  G.~Sporiš, ``A new method for assessing squash tactics using 15 court areas
  for ball locations,'' \emph{Human Movement Science}, vol.~34, pp. 81 -- 90,
  2014. [Online]. Available:
  \url{http://www.sciencedirect.com/science/article/pii/S0167945714000037}
\BIBentrySTDinterwordspacing

\bibitem{Tamaki_2013_CVPR_Workshops}
\BIBentryALTinterwordspacing
S.~Tamaki and H.~Saito, ``Reconstruction of 3d trajectories for performance
  analysis in table tennis,'' in \emph{{IEEE} Conference on Computer Vision and
  Pattern Recognition, {CVPR} Workshops 2013, Portland, OR, USA, June 23-28,
  2013}.\hskip 1em plus 0.5em minus 0.4em\relax {IEEE} Computer Society, 2013,
  pp. 1019--1026. [Online]. Available:
  \url{https://doi.org/10.1109/CVPRW.2013.148}
\BIBentrySTDinterwordspacing

\bibitem{Weeratunga_2017_CVPR_Workshops}
K.~Weeratunga, A.~Dharmaratne, and K.~B. How, ``Application of computer vision
  and vector space model for tactical movement classification in badminton,''
  in \emph{2017 IEEE Conference on Computer Vision and Pattern Recognition
  Workshops (CVPRW)}, July 2017, pp. 132--138.

\bibitem{ghosh2018smarttennistv}
A.~Ghosh and C.~Jawahar, ``Smarttennistv: Automatic indexing of tennis
  videos,'' in \emph{Computer Vision, Pattern Recognition, Image Processing,
  and Graphics: 6th National Conference, NCVPRIPG 2017, Mandi, India, December
  16-19, 2017, Revised Selected Papers 6}.\hskip 1em plus 0.5em minus
  0.4em\relax Springer, 2018.

\bibitem{he2017mask}
K.~He, G.~Gkioxari, P.~Doll{\'a}r, and R.~Girshick, ``Mask r-cnn,'' in
  \emph{Proceedings of the IEEE international conference on computer vision},
  2017, pp. 2961--2969.

\bibitem{lin2014microsoft}
T.-Y. Lin, M.~Maire, S.~Belongie, J.~Hays, P.~Perona, D.~Ramanan,
  P.~Doll{\'a}r, and C.~L. Zitnick, ``Microsoft coco: Common objects in
  context,'' in \emph{European conference on computer vision}.\hskip 1em plus
  0.5em minus 0.4em\relax Springer, 2014.

\bibitem{sun2018beyond}
Y.~Sun, L.~Zheng, Y.~Yang, Q.~Tian, and S.~Wang, ``Beyond part models: Person
  retrieval with refined part pooling (and a strong convolutional baseline),''
  in \emph{Proceedings of the European Conference on Computer Vision (ECCV)},
  2018.

\bibitem{lstm}
S.~Hochreiter and J.~Schmidhuber, ``Long short-term memory,'' \emph{Neural
  computation}, vol.~9, no.~8, pp. 1735--1780, 1997.

\bibitem{ullah2018action}
A.~Ullah, J.~Ahmad, K.~Muhammad, M.~Sajjad, and S.~W. Baik, ``Action
  recognition in video sequences using deep bi-directional lstm with cnn
  features,'' \emph{IEEE Access}, vol.~6, pp. 1155--1166, 2018.

\bibitem{simonyan2014very}
K.~Simonyan and A.~Zisserman, ``Very deep convolutional networks for
  large-scale image recognition,'' \emph{arXiv preprint arXiv:1409.1556}, 2014.

\bibitem{deng2009imagenet}
J.~Deng, W.~Dong, R.~Socher, L.-J. Li, K.~Li, and L.~Fei-Fei, ``Imagenet: A
  large-scale hierarchical image database,'' in \emph{Computer Vision and
  Pattern Recognition, 2009. CVPR 2009. IEEE Conference on}, 2009.

\bibitem{he2016deep}
K.~He, X.~Zhang, S.~Ren, and J.~Sun, ``Deep residual learning for image
  recognition,'' in \emph{Proceedings of the IEEE conference on computer vision
  and pattern recognition}, 2016, pp. 770--778.

\end{thebibliography}

\end{document}